\documentclass[9pt]{article}
\usepackage{spconf}
\usepackage{graphicx}
\usepackage{booktabs, multirow}
\usepackage{soul}
\usepackage{xcolor,colortbl}
\usepackage{changepage,threeparttable}
\usepackage{enumitem}
\usepackage{arydshln}
\usepackage{titlesec}
\usepackage{titlesec}
\titlespacing{\section}{0pt}{8pt plus 2pt minus 2pt}{4pt plus 1pt minus 1pt}
\titlespacing{\subsection}{0pt}{8pt plus 2pt minus 2pt}{4pt plus 1pt minus 1pt}

\title{2D REPRESENTATION FOR UNGUIDED SINGLE-VIEW 3D SUPER-RESOLUTION IN REAL-TIME}

\name{Ignasi Mas\textsuperscript{1}, Ivan Huerta, Ramon Morros, Javier Ruiz-Hidalgo}
\address{\textsuperscript{1}ignasi.mas.mendez@upc.edu}
\begin{document}
\ninept

\maketitle

\begin{abstract}
We introduce 2Dto3D-SR, a versatile framework for real-time single-view 3D super-resolution that eliminates the need for high-resolution RGB guidance. Our framework encodes 3D data from a single viewpoint into a structured 2D representation, enabling the direct application of existing 2D image super-resolution architectures. We utilize the Projected Normalized Coordinate Code (PNCC) to represent 3D geometry from a visible surface as a regular image, thereby circumventing the complexities of 3D point-based or RGB-guided methods. This design supports lightweight and fast models adaptable to various deployment environments. We evaluate 2Dto3D-SR with two implementations: one using Swin Transformers for high accuracy, and another using Vision Mamba for high efficiency. Experiments show the Swin Transformer model achieves state-of-the-art accuracy on standard benchmarks, while the Vision Mamba model delivers competitive results at real-time speeds. This establishes our geometry-guided pipeline as a surprisingly simple yet viable and practical solution for real-world scenarios, especially where high-resolution RGB data is inaccessible.
\end{abstract}
\begin{keywords}
3D, super-resolution, deep learning, real-time
\end{keywords}
\section{Introduction}
\label{sec:intro}

Single-view 3D Super-Resolution (SR) aims to reconstruct high-resolution geometry from a low-resolution observation, limited to a single viewpoint, such as a depth map or a point cloud (for the visible surface). This task is critical for applications in robotics and AR/VR.  Existing approaches typically fall into two main categories: point cloud upsampling methods (designed for any type of point cloud), such as PU-Net~\cite{pcu:punet}, and Depth Super-Resolution (DSR) techniques, like DJF~\cite{gdsr:djf}. The former operate directly on 3D data but often require complex processing or iterative refinement, limiting their suitability for real-time use. The latter tend to be less resource demanding but frequently depend on high-resolution (HR) RGB images to compensate for missing geometric context, introducing additional input dependencies and computational overhead.

To overcome these challenges, we propose \textbf{2Dto3D-SR}, a general framework that reframes single-view 3D SR as a 2D image-oriented problem. By encoding 3D geometry into a structured 2D representation, we can leverage the extensive progress and high efficiency of 2D image super-resolution (ISR) architectures. In this work, we demonstrate the potential of the framework using the Projected Normalized Coordinate Code (PNCC)~\cite{pncc:3ddfa} as our geometric representation. This approach eliminates the need for HR RGB guidance and avoids costly 3D operations. Our contributions are: (1) we propose a framework for unguided single-view 3D SR using 2D SR models; (2) we validate PNCC as an effective 2D representation for this task; and (3) we present two efficient implementations based on Swin Transformers and Vision Mamba that achieve state-of-the-art accuracy and real-time performance on standard DSR benchmarks, even without HR RGB input.

\section{Related Work}
\label{sec:relwork}

\noindent\textbf{3D to 2D Representation.} While traditional 2D formats for 3D data, such as depth or UV maps, are simple, they often discard geometric context. A more structured alternative is the Projected Normalized Coordinate Code (PNCC) from 3DDFA~\cite{pncc:3ddfa}, which encodes 3D coordinates into RGB images (detailed in Sec.~\ref{sec:3dto2d}). PNCC has been used for applications such as 3D portrait synthesis~\cite{pncc:real3dportrait} and fine-grained reconstruction~\cite{pncc:beyond3dmm}, but its use for single-view 3D super-resolution is, to our knowledge, novel.

\noindent\textbf{Image Super-Resolution (ISR).} ISR has evolved from early CNNs like SRCNN~\cite{2dsr:srcnn} and ESPCN~\cite{2dsr:espcn} to advanced models balancing fidelity and efficiency. SRGAN~\cite{2dsr:srgan} introduced perceptual realism, later refined by ESRGAN~\cite{2dsr:esrgan}. Attention and transformers further advanced the field, with RCAN~\cite{2dsr:rcan} using channel attention and SwinIR~\cite{2dsr:swinir} applying Swin Transformers. This was extended by SwinFIR~\cite{2dsr:swinfir} and HAT~\cite{2dsr:hat} with frequency modules and hybrid attention. While diffusion models like SRDiff~\cite{2dsr:srdiff} offer high quality at great cost, methods like ResShift~\cite{2dsr:resshift}, FSRDiff~\cite{2dsr:fsrdiff}, and AddSR~\cite{2dsr:addsr} improve efficiency. For lightweight applications, models like DRCT \cite{2dsr:drct}, HMANet~\cite{2dsr:hmanet}, CPAT~\cite{2dsr:cpat}, EDT~\cite{2dsr:edt}, and DVMSR~\cite{2dsr:dvmsr} focus on architectural optimizations or efficient training strategies (DVMSR uses Vision Mamba and is relevant to this work as we will use this architecture).

\noindent\textbf{Depth Super-Resolution (DSR).} Most DSR methods are \textit{guided}, enhancing Low Resolution (LR) depth with HR auxiliary inputs like RGB images. Early works extended joint filtering to deep learning, with DJF~\cite{gdsr:djf} and DJFN~\cite{gdsr:djfn} establishing CNN-based guidance, DMSG~\cite{gdsr:dmsg} exploring multiscale fusion, PAC~\cite{gdsr:pac} introducing pixel-adaptive convolutions, and DKN~\cite{gdsr:dkn} using deformable kernels. JIIF~\cite{gdsr:jiif} modeled depth as an implicit function for interpretability. More recent models use richer guidance: DCTNet~\cite{gdsr:dctnet} leverages frequency-domain learning, SUFT~\cite{gdsr:suft} employs uncertainty-aware feature transmission, and RSAG~\cite{gdsr:rsag} uses Structure Attention. Graphs have also been studied (\cite{gdsr:graphsr}). The current state-of-the-art, SGNet~\cite{gdsr:sgnet}, combines gradient and frequency cues. However, all these methods depend on well-aligned HR RGB data. In contrast, \textit{unguided} DSR (like CAIRL~\cite{udsr:cairl}) remains  under-explored. These methods, while still lagging in accuracy, are critical for cases where RGB is inaccessible, such as when a device is unavailable or the signal is omitted, thus motivating robust RGB-independent solutions.

\noindent\textbf{Point Cloud Upsampling (PCU).} PCU aims to create dense point clouds from sparse ones. PU-Net~\cite{pcu:punet} initiated deep PCU with local patches, while graph-based methods like PU-GCN~\cite{pcu:pugcn} and ARGCN~\cite{pcu:argcn} better captured global context, with the latter adding adversarial training. Other approaches include GANs with transformer-inspired features in PU-GAN~\cite{pcu:pugan}, gradient-based loss in Grad-PU~\cite{pcu:gradpu}, and ray-based priors in PU-Ray~\cite{pcu:puray}. RepKPU~\cite{pcu:repkpu} and APU-LDI~\cite{pcu:apuldi} explored improved representations and interpolation. Despite these advances, PCU methods are often computationally intensive and rely on iterative refinement, making them ill-suited for real-time applications.

\section{Methodology}
\label{sec:method}

\subsection{Pipeline Overview}
\label{sec:pipeline}
Our framework operates by converting input 3D data into a chosen 2D representation, where a 2D Super-Resolution (SR) model is applied. This representation must meet several criteria: it must encode purely geometric information to function without RGB or other auxiliary inputs; it must be reversible from a single viewpoint to allow lossless recovery; and it must retain an image-like structure for compatibility with standard 2D SR architectures. The SR model itself must be unguided, operating solely on LR representation.

\begin{figure*}[t]
  \centering
  \includegraphics[width=0.9\textwidth]{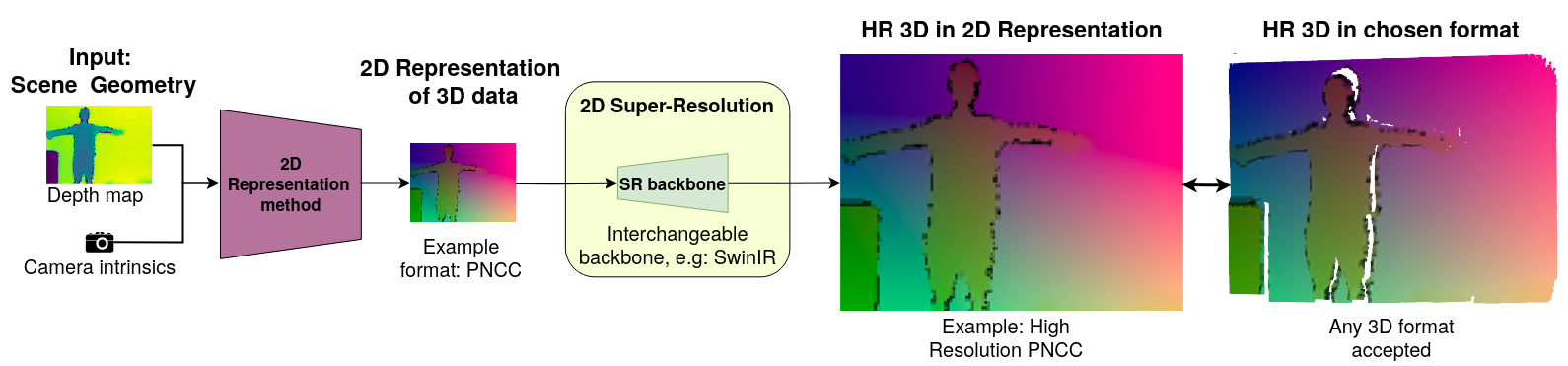}          
  \caption{Overview of our 2Dto3D-SR framework. A low-resolution depth map is first converted to a PNCC representation. A standard 2D SR model (here, SwinT-PNCC) is applied in this domain to produce a high-resolution PNCC map, which is then converted back to a high-resolution 3D representation like a depth map or point cloud.}
  \label{fig:pipeline-sv}
\end{figure*}

Figure~\ref{fig:pipeline-sv} illustrates this pipeline using PNCC as the 2D representation (detailed in Section~\ref{sec:3dto2d}). The example, using a sample from the RGB-D-D dataset, shows how an input depth map with intrinsic parameters is processed and ultimately reconstructed as a surface point cloud, demonstrating the framework's support for transforming the output back into any desired 3D format.

\subsection{PNCC as a 2D Representation}
\label{sec:3dto2d}
\begin{figure*}[t]
  \centering
  \includegraphics[width=0.67\textwidth]{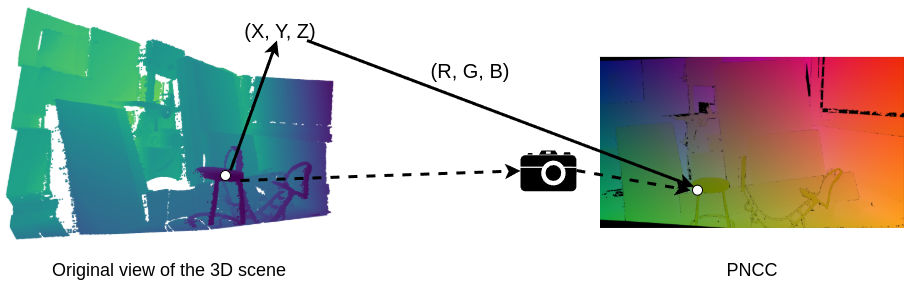}          
  \caption{Illustration of the PNCC representation of single-view 3D data.}
  \label{fig:pncccontent}
\end{figure*}

As discussed in Section~\ref{sec:pipeline}, selecting a suitable 3D representation is crucial. While depth maps lack spatial detail and point clouds are incompatible with 2D SR models, we adopt PNCC. Illustrated in Figure~\ref{fig:pncccontent}, PNCC encodes the normalized 3D coordinates $(X, Y, Z)$ of a scene point into the $(R, G, B)$ values of its corresponding pixel. This preserves full single-view geometric information in a format compatible with standard 2D architectures, making it a powerful choice for our framework. Crucially, PNCC is also reversible to both depth maps and point clouds.

Given a depth map $d(u,v)$ and camera intrinsics $(f_x, f_y, c_x, c_y)$, we compute PNCC as:
\begin{equation}
PNCC(u, v) = \left(\frac{(u - c_x) \cdot d(u, v)}{f_x \cdot s}, \frac{(v - c_y) \cdot d(u, v)}{f_y \cdot s}, \frac{d(u, v)}{s}\right)
\end{equation}
where $s$ is a scale factor. The coordinates are then normalized to fit the RGB range while preserving aspect ratio. Pixels with invalid depth are excluded from loss computation and evaluation, and are filled using nearest-neighbor interpolation for stability. This process is based on resolution-dependent camera intrinsics.

PNCC stores normalized XYZ coordinates for the visible surface. The Point Cloud is recovered by un-normalizing the coordinates of each valid pixel, skipping invalid ones. Similarly, the depth map is obtained by simply un-normalizing the blue channel (Z).

\subsection{2D Super-Resolution Architectures}
We adapt two leading 2D ISR models for our framework:
\begin{itemize}
    \item \textbf{SwinT-PNCC}: Based on the high-accuracy SwinIR~\cite{2dsr:swinir} architecture, this model uses a Swin Transformer backbone for deep feature extraction.
    \item \textbf{VM-PNCC}: Based on the highly efficient DVMSR~\cite{2dsr:dvmsr}, uses a Vision Mamba backbone to achieve real-time speeds.
\end{itemize}

The SwinIR and DVMSR architectures, detailed in their respective papers, both use low-level and high-level (backbone) feature extractors, followed by an upsampling CNN that fuses these features. We implement two variants of each strategy, both using PNCC as input: one predicts the complete PNCC (XYZ), while the other predicts only the Z channel (depth), allowing subsequent PNCC computation as explained in~\ref{sec:3dto2d}. We ablate this choice in Section~\ref{sec:results} (\textit{Predicting XYZ vs predicting only Z}).

Both models are trained from scratch directly on PNCC data using a pixel-wise Charbonnier loss.

\section{Experiments and Results}
\label{sec:experiments_results}

\subsection{Experimental Setup}
We evaluate our models against a bicubic baseline and several state-of-the-art guided methods. For context, we compare against the reported results of \textbf{GraphSR}~\cite{gdsr:graphsr} and \textbf{DCTNet}~\cite{gdsr:dctnet} on the standard NYUv2 benchmark (using the SGNet preprocessed version to do so). For direct and controlled comparisons across all our test sets, we select \textbf{SGNet}~\cite{gdsr:sgnet} as our primary baseline, as it represents the best-performing guided method. For fairness, SGNet and our models are all trained on our custom-processed, geometrically-aligned version of the NYUv2 dataset~\cite{data:nyuv2} (which allows us additional qualitative comparison, as the SGNet dataset version does no align well when visualized). The upcoming paragraph explains the process used to construct this dataset. We test generalization on the Middlebury~\cite{data:middlebury} and real-world RGB-D-D~\cite{data:rgbdd} datasets. Although we initially consider the Lu dataset~\cite{data:lu}, it is excluded due to missing parameters and limited contribution beyond Middlebury. Performance is measured by RMSE (cm) on valid depth regions at $\times4$, $\times8$, and $\times16$ scales. We measure the effectiveness by the inference time and number of parameters (all tests under NVIDIA RTX a6000 GPU).

For the (well aligned) NYUv2 and Middlebury datasets, we generate low-resolution (LR) inputs by bicubically downsampling the raw depth, with corresponding valid masks created by min-pooling the HR masks to ensure spatial consistency (see Section~\ref{sec:3dto2d}). The RGB-D-D dataset, which already includes real-world HR and LR captures from different devices, presents a realistic domain gap. For this dataset, we construct  $\times4$ pairs via additional bicubic downscaling to match our target resolution ratios. All models are trained on the correctly aligned NYUv2 set and evaluated on the NYUv2 test split, Middlebury, and RGB-D-D. To generate the PNCCs, we use the method explained in Section~\ref{sec:3dto2d} with the corresponding intrinsics for each dataset.

\subsection{Results}
\label{sec:results}

%Please add the following packages if necessary:
%\usepackage{booktabs, multirow} % for borders and merged ranges
%\usepackage{soul}% for underlines
%\usepackage{xcolor,colortbl} % for cell colors
%\usepackage{changepage,threeparttable} % for wide tables
%If the table is too wide, replace \begin{table}[!htp]...\end{table} with
%\begin{adjustwidth}{-2.5 cm}{-2.5 cm}\centering\begin{threeparttable}[!htb]...\end{threeparttable}\end{adjustwidth}

\setlength{\tabcolsep}{3pt}

While point cloud upsampling methods like Grad-PU~\cite{pcu:gradpu} are a natural choice for 3D super-resolution, they prove unsuitable for single-view scenarios. Despite its compact size, Grad-PU suffers from limited performance and long runtimes due to its iterative nature (see Table~\ref{tab:pcu}). As such, we exclude point-based methods and focus on image-based Depth Super-Resolution.

\begin{table}[!htb]
\centering
\begin{threeparttable}
\scriptsize
\begin{tabular}{p{3.5cm}|r|r|r}
\toprule
\textbf{APPROACH} & 
\textbf{CHAMFER} & 
\textbf{TIME (s)} & 
\textbf{\# PARAMS} \\
&\textbf{DISTANCE} & & \\
\toprule
SwinT-PNCC & \textbf{0.163} & \textbf{0.164} & 11.7M \\
Grad-PU    & 0.420          & 9.920 & \textbf{67K} \\
\bottomrule
\end{tabular}
\caption{Point Cloud Upsampling results in NYUv2 $\times$4}
\label{tab:pcu}
\end{threeparttable}
\end{table}

On the incorrectly aligned version of the NYUv2 benchmark (Table~\ref{tab:resnyuv2sgnet}), although our models do not achieve the same level of accuracy as SGNet and DCTNet, they operate with significantly reduced runtime and size (they also outperform GraphSR). On the well aligned version (Table~\ref{tab:datasets_results}), they outperform SGNet at $\times4$ and $\times8$, and remain competitive at $\times16$, while being up to $\times18$ faster. Generalization is further validated on Middlebury, where SGNet degrades at higher scales, even falling below bicubic. In contrast, our models maintain strong performance across all scales. On the challenging RGB-D-D dataset, our method clearly surpasses both SGNet and bicubic, despite sensor noise and natural degradation. In all the tables, best values on each experiment are bolded. Bicubic is included to the tables but excluded from this comparison.

\begin{table}[!htb]
\centering
\begin{threeparttable}
\label{tab:resnyuv2sgnet}
\begin{tabular}{l|c|c|c}
\toprule
\textbf{Method} & \textbf{RMSE (cm) $\downarrow$} & \textbf{Time (s) $\downarrow$} & \textbf{\# Params} \\
\midrule
GraphSR (Guided)\tnote{*} & 2.54 & - & - \\
DCTNet (Guided)\tnote{*} & 1.59 & - & - \\
\midrule
SGNet (Guided) & \textbf{1.16} & 0.59  & 36.4M \\
SwinT-PNCC & 2.19          & \textbf{0.16}  & \textbf{11.7M} \\
\bottomrule
\end{tabular}
\begin{tablenotes}
    \item[*] Results reported from original paper~\cite{gdsr:graphsr, gdsr:dctnet}.
\end{tablenotes}
\caption{Comparison on NYUv2 (SGNet version) at $\times$4.}
\end{threeparttable}
\end{table}

\begin{table}[!htb]
\centering
\footnotesize
\setlength{\tabcolsep}{3pt} % Adjust column spacing
\begin{threeparttable}
\label{tab:datasets_results}
\begin{tabular}{l|ccc|ccc|c}
\toprule
& \multicolumn{3}{c|}{\textbf{NYUv2 (aligned)}} & \multicolumn{3}{c|}{\textbf{Middlebury}} & \textbf{RGB-D-D} \\
\cmidrule(lr){2-4} \cmidrule(lr){5-7} \cmidrule(lr){8-8}
\textbf{APPROACH} & \textbf{x4} & \textbf{x8} & \textbf{x16} & \textbf{x4} & \textbf{x8} & \textbf{x16} & \textbf{x4} \\
\midrule
\multicolumn{8}{l}{\textit{RMSE (cm) $\downarrow$}} \\
\midrule
Bicubic      & 25.22 & 34.80 & 47.19 & 18.27 & 25.60 & 70.52 & 0.222 \\
SGNet        & 11.66 & 21.26 & \textbf{35.85} & 14.93 & 40.38 & 73.98 & 0.232 \\
SwinT-PNCC   & \textbf{9.99}  & \textbf{19.15} & 39.04 & \textbf{7.91}  & \textbf{20.52} & \textbf{37.42} & \textbf{0.212} \\
VM-PNCC      & 11.19 & 20.15 & 39.48 & 11.95 & 22.97 & 37.81 & - \\
\midrule
\multicolumn{8}{l}{\textit{TIME (s) $\downarrow$}} \\
\midrule
Bicubic      & 0.002 & 0.002 & 0.002 & 0.004 & 0.002 & 0.002 & 0.002 \\
SGNet        & 0.596 & 0.438 & 0.436 & 0.671 & 0.498 & 0.501 & 0.381 \\
SwinT-PNCC   & 0.164 & 0.121 & 0.055 & 0.342 & 0.131 & 0.093 & 0.109 \\
VM-PNCC      & \textbf{0.045} & \textbf{0.027} & \textbf{0.024} & \textbf{0.061} & \textbf{0.074} & \textbf{0.047} & - \\
\midrule
\multicolumn{8}{l}{\textit{\# PARAMS}} \\
\midrule
SGNet        & 36.4M & 39.9M & 86.6M & 36.4M & 39.9M & 86.6M & 36.4M \\
SwinT-PNCC   & 11.7M & 11.8M & 11.8M & 11.7M & 11.8M & 11.8M & 11.7M \\
VM-PNCC      & \textbf{7.2M}  & \textbf{7.3M}  & \textbf{7.3M}  & \textbf{7.2M}  & \textbf{7.3M}  & \textbf{7.3M}  & - \\
\bottomrule
\end{tabular}
\caption{Comparison across several scales at the NYUv2, Middlebury, and RGB-D-D datasets. All models trained on NYUv2 train set.}
\end{threeparttable}
\end{table}

Overall, our models achieve state-of-the-art results in unguided Depth Super-Resolution, while being significantly faster, smaller, and more robust across datasets and real-world conditions.

We provide qualitative comparisons between SwinT-PNCC, SGNet, and bicubic upsampling to support our quantitative results.

\begin{figure*}[!htb]
\centering
\begin{minipage}[b]{0.15\textwidth}
  \centering
  \includegraphics[width=\textwidth]{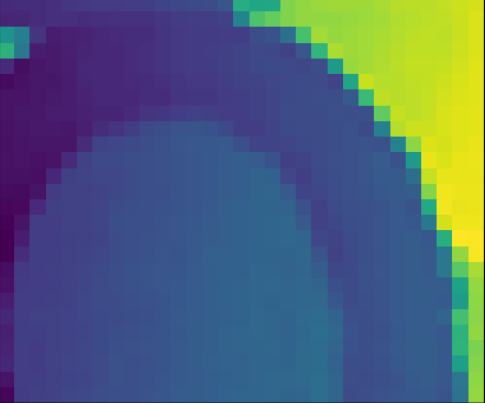}
  \\ \small Input
\end{minipage}
\begin{minipage}[b]{0.15\textwidth}
  \centering
  \includegraphics[width=\textwidth]{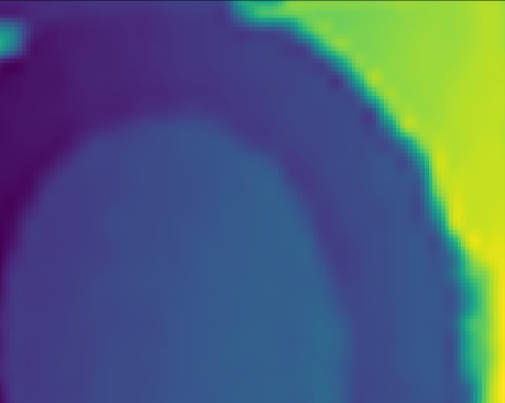}
  \\ \small Bicubic
\end{minipage}
\begin{minipage}[b]{0.15\textwidth}
  \centering
  \includegraphics[width=\textwidth]{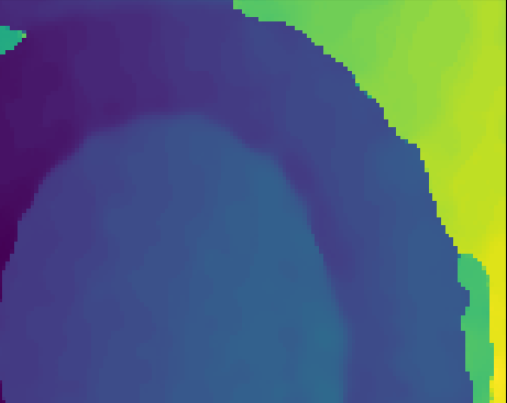}
  \\ \small SGNet
\end{minipage}
\begin{minipage}[b]{0.15\textwidth}
  \centering
  \includegraphics[width=\textwidth]{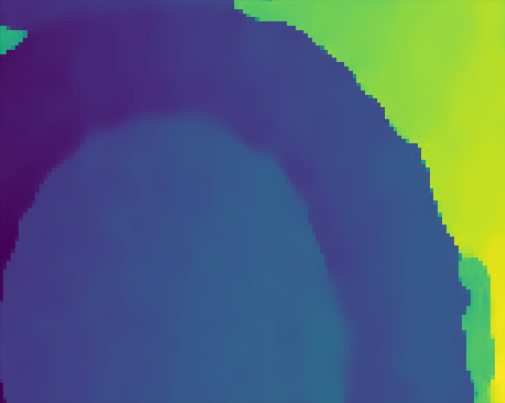}
  \\ \small SwinT-PNCC
\end{minipage}
\begin{minipage}[b]{0.15\textwidth}
  \centering
  \includegraphics[width=\textwidth]{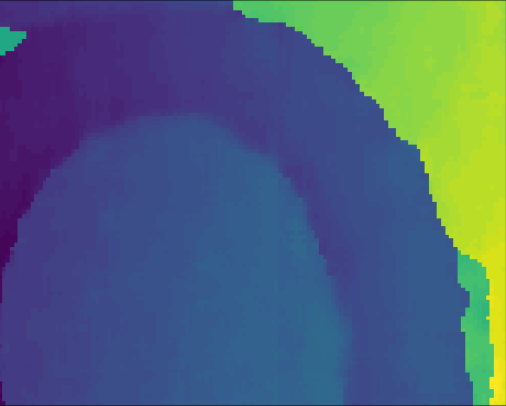}
  \\ \small Target
\end{minipage}

\caption{Qualitative Results: Predicted depth of different experiments in aligned NYUv2 at $\times$4}
\label{fig:qualitative_pncc_row}
\end{figure*}

Figure~\ref{fig:qualitative_pncc_row} shows the predicted depth map visualizations on a NYUv2 sample with a $\times$4 upscaling factor. In this example the superiority of learning-based methods (SGNet, SwinT-PNCC) over traditional techniques (bicubic) is clear, particularly at the edges, where bicubic displays blurriness.

\begin{figure*}[!htb]
\centering

% 1. INPUT
\begin{minipage}[b]{0.17\textwidth}
  \centering
  \includegraphics[width=\textwidth]{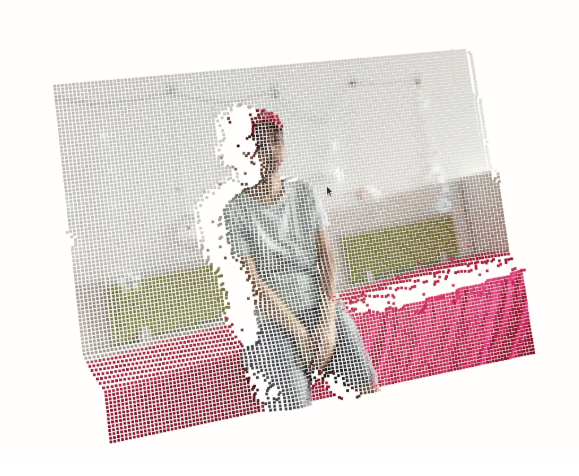} \\[1mm]
  \includegraphics[width=\textwidth]{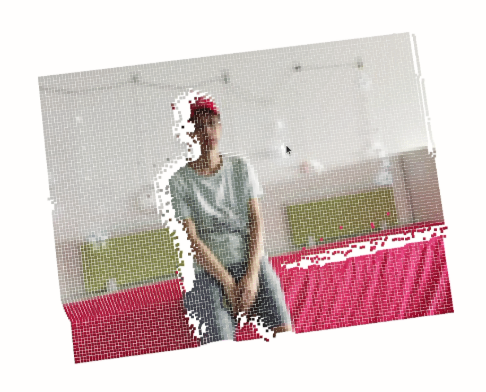}
  \\ \small Input
\end{minipage}
% 2. BICUBIC
\begin{minipage}[b]{0.17\textwidth}
  \centering
  \includegraphics[width=\textwidth]{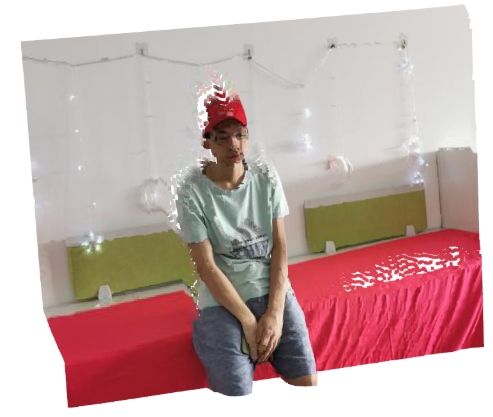} \\[1mm]
  \includegraphics[width=\textwidth, trim=20mm 0mm 20mm 40mm, clip]{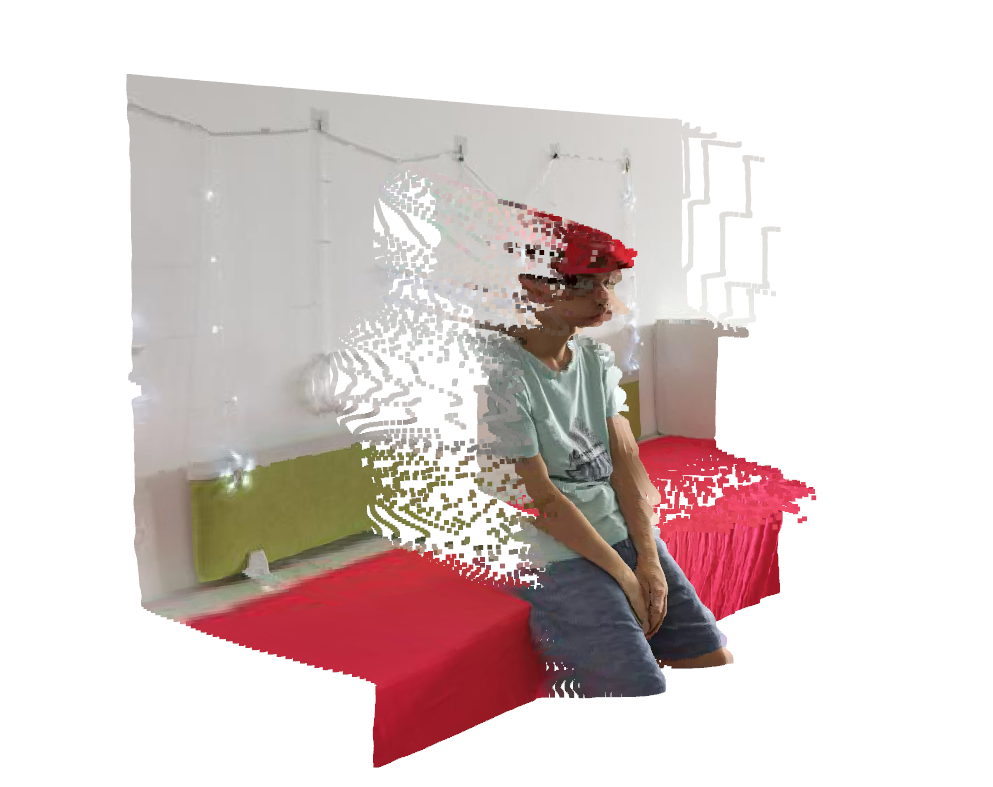}
  \\ \small Bicubic
\end{minipage}
% 3. SGNET
\begin{minipage}[b]{0.17\textwidth}
  \centering
  \includegraphics[width=\textwidth, trim=60mm 0mm 60mm 0mm, clip]{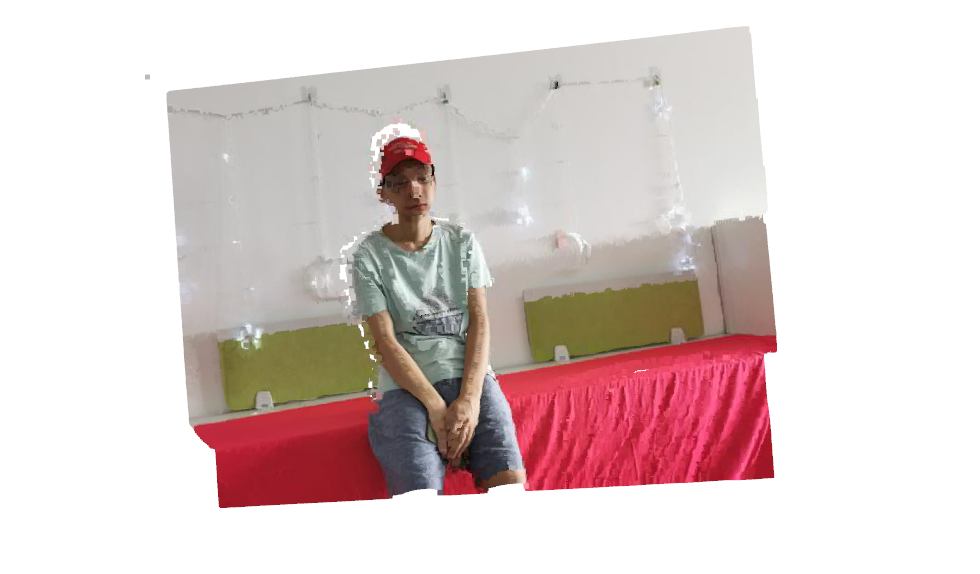} \\[1mm]
  \includegraphics[width=\textwidth, trim=60mm 0mm 60mm 40mm, clip]{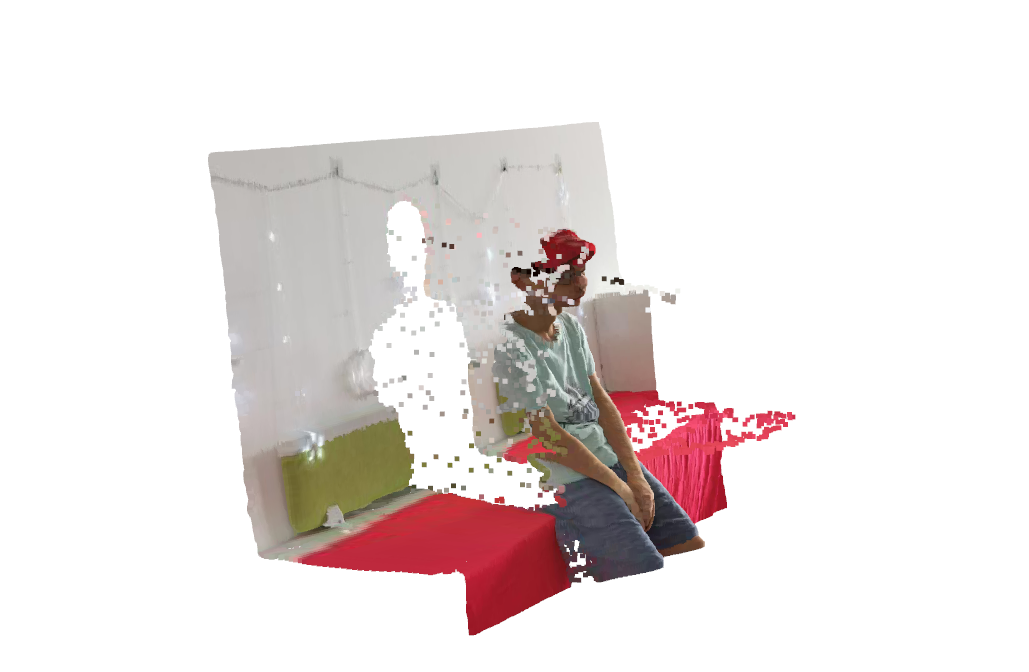}
  \\ \small SGNet
\end{minipage}
% 4. SWINT-PNCC
\begin{minipage}[b]{0.17\textwidth}
  \centering
  \includegraphics[width=\textwidth, trim=90mm -40mm 90mm 0mm, clip]{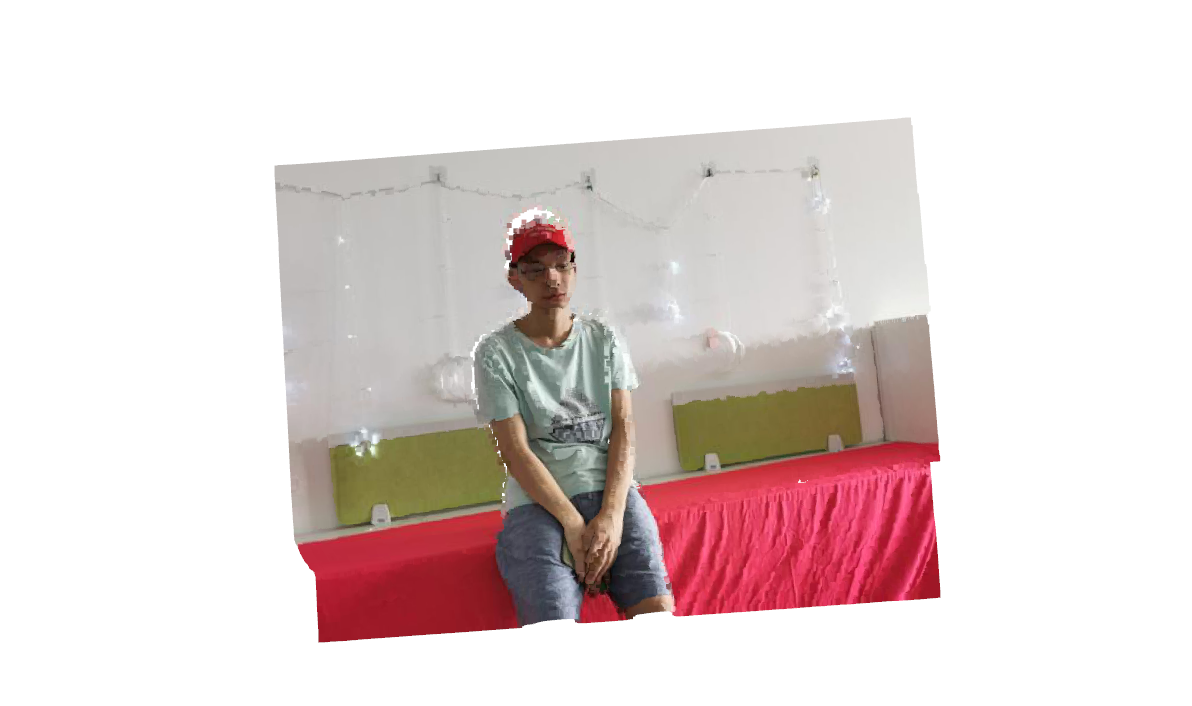} \\[1mm]
  \includegraphics[width=\textwidth, trim=40mm 0mm 140mm 70mm, clip]{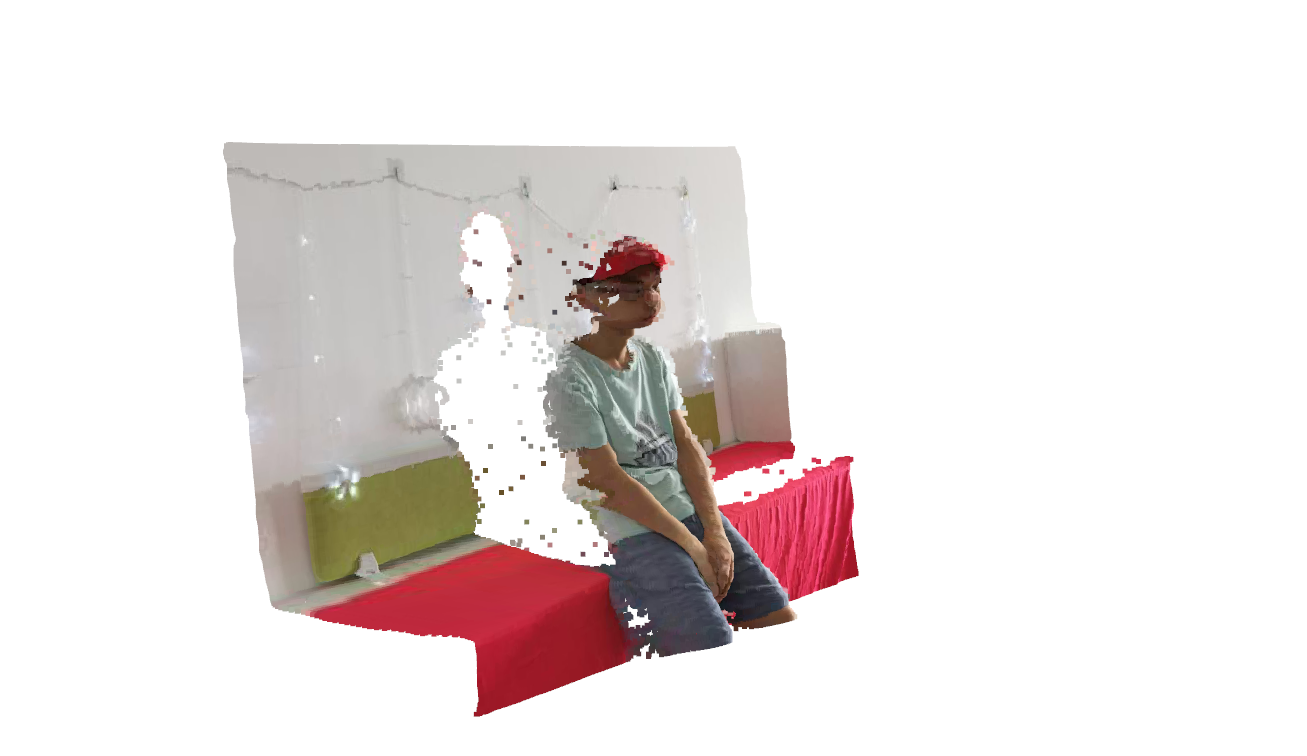}
  \\ \small SwinT-PNCC
\end{minipage}
% 5. TARGET
\begin{minipage}[b]{0.17\textwidth}
  \centering
  \includegraphics[width=\textwidth, trim=45mm -20mm 45mm 0mm, clip]{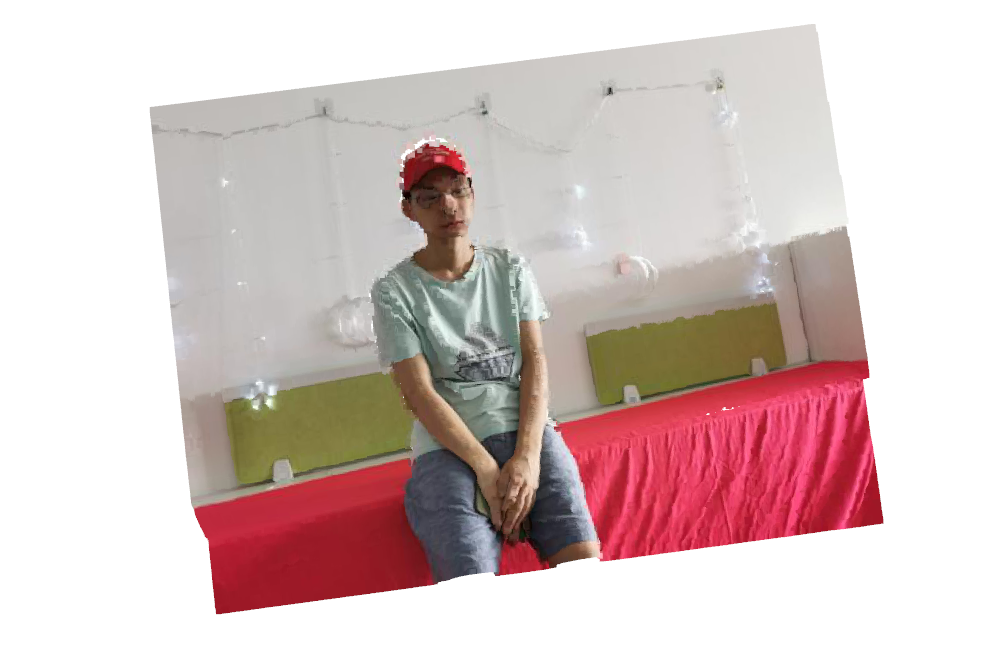} \\[1mm]
  \includegraphics[width=\textwidth, trim=20mm 0mm 20mm 10mm, clip]{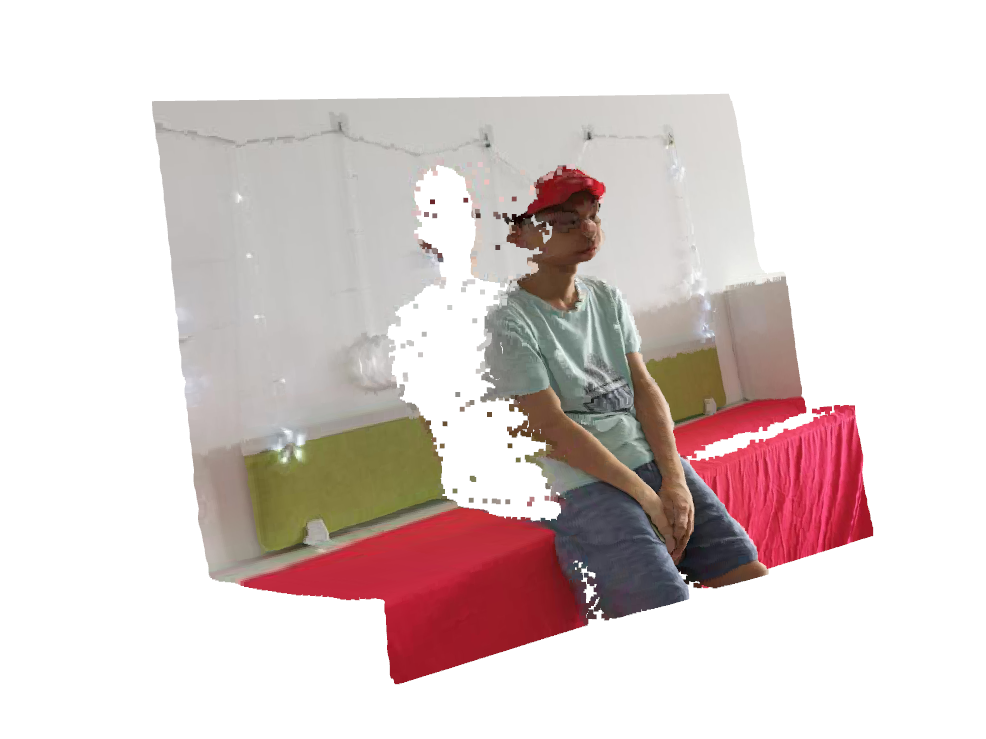}
  \\ \small Target
\end{minipage}

\vspace{1mm}
\caption{Qualitative Results: Single-view Point Clouds across different methods (2 views of the surface for each) on RGB-D-D at $\times$4. RGB is added for visualization to all methods, including the unguided ones. Noisy hulls are observed with bicubic upscaling, while other methods exhibit scattered outliers around the scene, noticeable near the person's cap and around the bed and more present in SGNet than SwinT-PNCC.}
\label{fig:qualitative_pc}
\end{figure*}

To better evaluate spatial accuracy, Figure~\ref{fig:qualitative_pc} presents two 3D views of the reconstructed point clouds from RGB-D-D at the same scale. Depth maps are backprojected using intrinsics, and HR RGB is used only for rendering. SwinT-PNCC produces the most coherent geometry, with smooth surfaces and accurate structures. SGNet improves over bicubic but introduces noise artifacts that our model avoids (the outlier points around the cap of the SGNet result are a clear example), highlighting its superior realism and consistency.

\textbf{Ablation study: Predicting XYZ vs predicting only Z.} We compare two output strategies for reconstructing 3D geometry from PNCC input. The first predicts all three PNCC channels (XYZ), while the second predicts only the Z component (depth), using projection techniques (section~\ref{sec:3dto2d}) with known intrinsics to recover the full geometry. The latter enforces geometric consistency by constraining outputs to camera rays. As shown in Table~\ref{tab:xyz_vs_z}, predicting depth yields comparable or slightly better RMSE, likely due to improved alignment with scene geometry.

\begin{table}[!htb]
\centering
\begin{threeparttable}
\begin{tabular}{p{3cm}p{3cm}|r|r}
\toprule
\textbf{TEST SET} & \textbf{APPROACH} & \multicolumn{2}{c}{\textbf{RMSE (cm)}} \\
\toprule
 & & \textbf{Z} & \textbf{XYZ} \\
\cmidrule{0-2}\cmidrule{3-4}
\multirow{2}{*}{NYUv2}       & SwinT-PNCC & \textbf{9.99}  & 10.47 \\
                             & VM-PNCC    & \textbf{11.19} & 11.39 \\
\midrule
\multirow{2}{*}{Middlebury}  & SwinT-PNCC & \textbf{7.91}  & 9.26 \\
                             & VM-PNCC    & \textbf{11.95} & 12.06 \\
\midrule
RGB-D-D                      & SwinT-PNCC & 0.212          & \textbf{0.212} \\
\bottomrule
\end{tabular}
\caption{RMSE (cm) at scale $\times$4 across datasets comparing \textbf{SwinT-PNCC} and \textbf{VM-PNCC} using either Z or XYZ outputs. Best values are bolded.}
\label{tab:xyz_vs_z}
\end{threeparttable}
\end{table}

\textbf{Ablation study: Depth as input.} To evaluate the improvement of using the PNCC representation in our method, we performed a comparison between the VM-Depth approach using depth maps as input and the standard approach employing the PNCC on our aligned NYUv2 dataset with x4 upscaling. The comparison results are detailed in Table~\ref{tab:deptthinput}. This highlights the effectiveness of the PNCC representation over simple Depth maps for single-view 3D SR, with the alteration of input alone resulting in improvement.

\begin{table}[!htb]
\centering
\begin{threeparttable}
\begin{tabular}{p{2cm}p{1cm}|r|r|r}
\toprule
\textbf{APPROACH} & \textbf{INPUT} & \textbf{RMSE (cm)} & \textbf{TIME (s)} & \textbf{\# PARAMS} \\
\toprule
VM-Depth & Depth & 11.34 & \textbf{0.043} & \textbf{7.2M} \\
VM-PNCC  & PNCC & \textbf{11.19} & 0.045 & 7.2M \\
\bottomrule
\end{tabular}
\caption{Our VM-PNCC approach using a Depth representation vs using PNCC. Trained and evaluated in NYUv2 (different sets).}
\label{tab:deptthinput}
\end{threeparttable}
\end{table}

\section{Conclusion}
\label{sec:conclusion}

In this work, we introduced \textbf{2Dto3D-SR}, a general framework for single-view 3D super-resolution that leverages 2D representations of 3D geometry. We validated the effectiveness of PNCC as a compact and reversible encoding, implementing two variants, \textbf{SwinT-PNCC} (Swin Transformer) and \textbf{VM-PNCC} (Vision Mamba), to showcase the framework's flexibility. Our models achieve state-of-the-art results in accuracy and efficiency across diverse benchmarks without relying on RGB guidance, offering an efficient bridge between 2D models and 3D data enhancement. While the current design for a single view can be a limitation in some applications, the framework can be extended to a multi-view setting. Future work will focus in that direction as well as applying other types of 3D data enhancement.

\bibliographystyle{IEEEbib}
\bibliography{ref}

\end{document}